%% file: main.tex
\documentclass{article} % For LaTeX2e
\usepackage{iclr2020_conference,times}

\input{math_commands.tex}

\usepackage[hidelinks]{hyperref}
\usepackage{url}
\usepackage{microtype}

\usepackage{graphicx}
\usepackage{xcolor}
\usepackage{caption}
\usepackage{float}

\title{Analyzing Autoencoder-based Acoustic Word Embeddings}

\author{Yevgen Matusevych \\
School of Informatics\\
University of Edinburgh\\
\texttt{ymatusev@ed.ac.uk} \\
\And
Herman Kamper \\
E\&E Engineering \\
Stellenbosch University \\
\texttt{kamperh@sun.ac.za} \\
\And
Sharon Goldwater \\
School of Informatics\\
University of Edinburgh\\
\texttt{sgwater@inf.ed.ac.uk}
}

\iclrfinalcopy % Uncomment for camera-ready version, but NOT for submission.
\begin{document}

\maketitle

\begin{abstract}
Recent studies have introduced methods for learning acoustic word embeddings (AWEs)---fixed-size vector representations of words which encode their acoustic features. Despite the widespread use of AWEs in speech processing research, they have only been evaluated quantitatively in their ability to discriminate between whole word tokens.
% The characteristics of the learned representations are therefore only weakly understood.
To better understand the applications of AWEs in various downstream tasks and in cognitive modeling, we need to analyze the representation spaces of AWEs.
Here we analyze basic properties of AWE spaces learned by a sequence-to-sequence encoder-decoder model in six typologically diverse languages. We first show that these AWEs preserve some information about words' absolute duration and speaker. At the same time, the representation space of these AWEs is organized such that the distance between words' embeddings increases with those words' phonetic dissimilarity. Finally, the AWEs exhibit a word onset bias, similar to patterns reported in various studies on human speech processing and lexical access. We argue this is a promising result and encourage further evaluation of AWEs as a potentially useful tool in cognitive science, which could provide a link between speech processing and lexical memory.
\end{abstract}

\section{Introduction}

% In many areas of natural language processing, lexical semantics, language cognition, etc., distributional word embeddings have become a de facto standard for representing word meanings \citep{}. Their ubiquitous use inspired a whole mini-field of studies analyzing their properties \citep{}. 

Several recent studies have introduced acoustic word embeddings (AWEs). AWEs are vector representations of individual word tokens based on their acoustic features \citep[][etc.]{levin2013, chung2016, holzenberger2018, kamper2019} rather than on their relation to other words, as in semantic (textual) word embeddings.\footnote{Although see \citet[][]{chung2018a, chung2018b} for speech-based \textit{semantic} embeddings, which we do not consider here.} Acoustic words unfold dynamically in time and have variable duration, yet fixed-dimensional AWEs have shown good performance in speech processing tasks such as word discrimination, and a recent study also suggests they can correctly predict some patterns of infant phonetic learning \citep{matusevych2020}. These results encourage exploration of AWEs for cognitive modeling, just as semantic word embeddings have been used as models of human semantic memory \citep[e.g.,][]{grand2018, nematzadeh2017, pereira2016}. As a first step, we need to describe the basic properties of AWEs and compare them to patterns observed in human lexical memory and spoken word perception, to better understand how temporal sequences of phones are encoded into static holistic representations, and whether these representations might correspond to human lexical representations. To our knowledge, one existing study \citep{ghannay2016} evaluates properties of AWEs, but it focuses on comparing them to orthographic word embeddings and only considers one language, French.

In this study, we consider AWEs in six different languages generated by a recent speech representation learning model, a correspondence-autoencoding recurrent neural network \citep[CAE-RNN;][]{kamper2019}, and analyze their basic properties to understand the organizing principles of the AWE space. An acoustic word contains three types of signal: (i)~properties specific to the particular instance of the word (in this study, we focus on one such feature, absolute duration), (ii)~the speakers' characteristics (i.e., all acoustic words spoken by the same person share some acoustic properties), and (iii)~the word's phonetic properties (i.e., \textit{cat} is more similar to \textit{catch} than to \textit{dog}). Like many other AWE models, the CAE-RNN is designed to abstract away from the first two types of information and learn the similarities between various spoken instances of the same word, similar to spoken word recognition in human speakers, who can identify the wordform (lexical item) regardless of who pronounces it and how. 
% Existing work shows that the CAE-RNN succeeds in doing this: it can discriminate between pairs of same vs.\ different words and cluster together different instances of the same word in its embedding space \citep{kamper2019, kamper2020}.
% % We test whether the AWEs encode the first two types of information, and whether their space is organized to prioritize the third type---i.e., word-invariant information.
% Here, we show that AWEs generated by the CAE-RNN encode some information about an acoustic word's absolute duration (type i) and speaker identity (type ii), although they abstract away from these two kinds of information (compared to a simple baseline derived from data) and succeed in learning the words' phonetic content (type iii).
% We show that AWEs generated by the CAE-RNN (1) encode information about the duration of a word and the number of phones in it, (2) abstract away from speaker characteristics and shape around word phonetic similarity and word identity, and (3) show the word onset bias often reported in literature on human speech processing and lexical access---i.e., human speakers attending more to initial sounds of a word than to other sounds \citep[][etc.]{gow1996, fougeron1997, keating1999}. 
%As a small case study, 
Existing work shows that the CAE-RNN succeeds in doing this: relative to a baseline that uses traditional signal processing methods, it is better at discriminating between pairs of same vs.\ different words and at clustering together different instances of the same word in its embedding space \citep{kamper2019, kamper2020}. 
%Nevertheless,
At the same time,
we show here that AWEs generated by the CAE-RNN
%still encode some 
do not completely abstract away from the
information about an acoustic word's absolute duration and speaker identity. More interestingly from a cognitive perspective, we also demonstrate that the AWEs exhibit a word onset bias, corresponding to a broad range of patterns reported in literature on human speech processing and lexical access which suggest that humans consider the initial sound of the word more `prominent' than its other sounds: for example, speakers emphasize it in articulation \citep{fougeron1997, keating1999}, listeners can capture the distinctions between word-initial and word-final sounds \citep{shatzman2006}, initial sounds have a special status in spoken word recognition \citep{marslen1989}, and the first letter is a more efficient cue for lexical retrieval than other letters \citep{brown1990}.

\section{Method}

The CAE-RNN model \citep[][]{kamper2019}, which we use for obtaining AWEs, is inspired by a sequence-to-sequence autoencoder, in which both the encoder and the decoder are RNNs \citep{chung2016}. During training, the CAE-RNN receives two different instances of the same wordform at a time: it encodes one of them into a vector of fixed dimensionality and uses this vector to reconstruct the other instance sequentially. Each instance is represented as a sequence of  \textit{frames}, where a frame is a 13-dimensional vector of mel-frequency cepstral coefficients (a standard way of representing the energy spectrum) extracted from a 25-ms-long slice of speech. Learning the correspondence between two instances of the same word encourages the model to abstract away from random noise and speaker characteristics while learning to encode the word-invariant phonetic information. This top-down guidance from the word level finds parallels in studies showing that even 6--8-month infants can  recognize some wordforms in running speech \citep[e.g.,][]{jus95a,jus99b}, and that this information can be useful for learning phonetic information \citep{feldman2013}.

% Following \citet[][]{kamper2019, kamper2020}, we train a set of models on a random sample of $100,000$ ground truth pairs of the same word type (obtained through forced word alignments) from six typologically diverse languages (see Appendix) in GlobalPhone, a non-parallel corpus of read newspaper articles \citep{schultz2002}, with $16$ hours of training and $2$ hours of test data on average. We adopt the existing architecture: $3$ hidden layers ($400$ gated recurrent units each) in both the decoder and encoder, and an embedding dimensionality of $130$.\footnote{On the `same-different' task, these models score $60$--$85\%$, depending on the language \citep[][]{kamper2020}.} Using the encoder of each model, we obtain AWEs for a set of unseen test words in the corresponding language.
% \ym{Put more technical details?}
% It is trained either on ground truth pairs (GT) from six GlobalPhone languages or on word pairs discovered automatically using a method of unsupervised term discovery (UTD) (REF). We additionally consider a so-called downsampling baseline, which creates a word's acoustic embedding by sampling its features, thus ensuring that the dimensionality is fixed (REF).
Following \citet[][]{kamper2020}, we train a set of models on six typologically diverse languages (see Appendix for details) from the GlobalPhone corpus \citep{schultz2002}. Using the encoder of each model, we obtain AWEs for a set of unseen test words in the corresponding language.
On these AWEs, we run a series of tests focusing on three main questions: (1) Do these AWEs preserve some information about speaker characteristics and segment acoustic properties (namely, its duration)? (2) Can AWEs abstract away from these two types of signal in favor of linguistically meaningful information, such as word phonetic similarity? (3) Can AWEs exhibit the human-like word onset bias? To address these questions, we probe the structure of the AWEs using three methods: (i)~using linear
classifiers\footnote{Linear classifiers allow for making claims about linear separability of the classes in an embedding space, a finding much easier to interpret than a potentially high performance of a complex nonlinear classifier.}
or regressions trained on top of AWEs; (ii)~using a machine ABX task \citep{schatz2013}, in which the distance between words A and X is compared to the distance between words B and X; and (iii)~directly comparing the distances between pairs of words meeting specific criteria.

To see how much the results rely on representation learning, we compare to a simple \emph{downsampling} baseline \citep[DS;][]{holzenberger2018}, which creates 130-dimensional embeddings (the same as the CAE-RNN AWEs) by concatenating 10 frames from the input word, equally spaced in time. 
%As a simple baseline, we consider downsampling \citep[DS,][]{holzenberger2018}: concatenating a number of frames equally spaced in time in a word, so that all words are represented as 130-dimensional vectors, similar to the CAE-RNN AWEs. Comparing the AWEs to this baseline tells us whether a particular observed property is a feature of the embedding space learned by the model, or this property is directly available in the kind of data that the model is trained on.

% The last method is more qualitative. It involves taking a set of word tokens, embedding these, and then calculating the (cosine) distance between all pairs of words in the embedding space. Box plots (for instance) are then used to compare the distances of words meeting different criteria, e.g.\ comparing the distances of words spoken by the same speaker to those spoken by a different speaker.

\section{Experiments and results}

% \subsection{Acoustic characteristics}

\begin{figure}
    \begin{minipage}{0.32\textwidth}
         \centering
         \includegraphics[width=\textwidth]{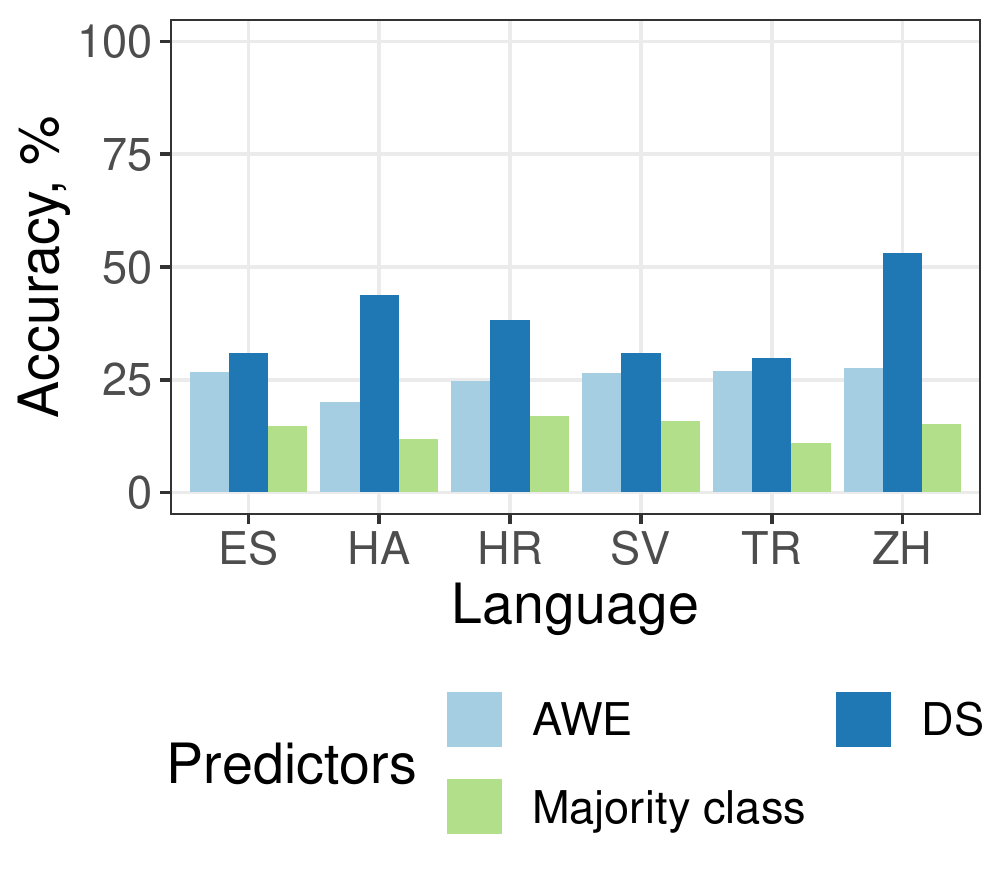}
        \caption{Classification accuracy of models predicting a word's speaker identity.}
        \label{fig:spkid}
    \end{minipage}\hfill
    \begin{minipage}{0.32\textwidth}
        \centering
        \includegraphics[width=\linewidth]{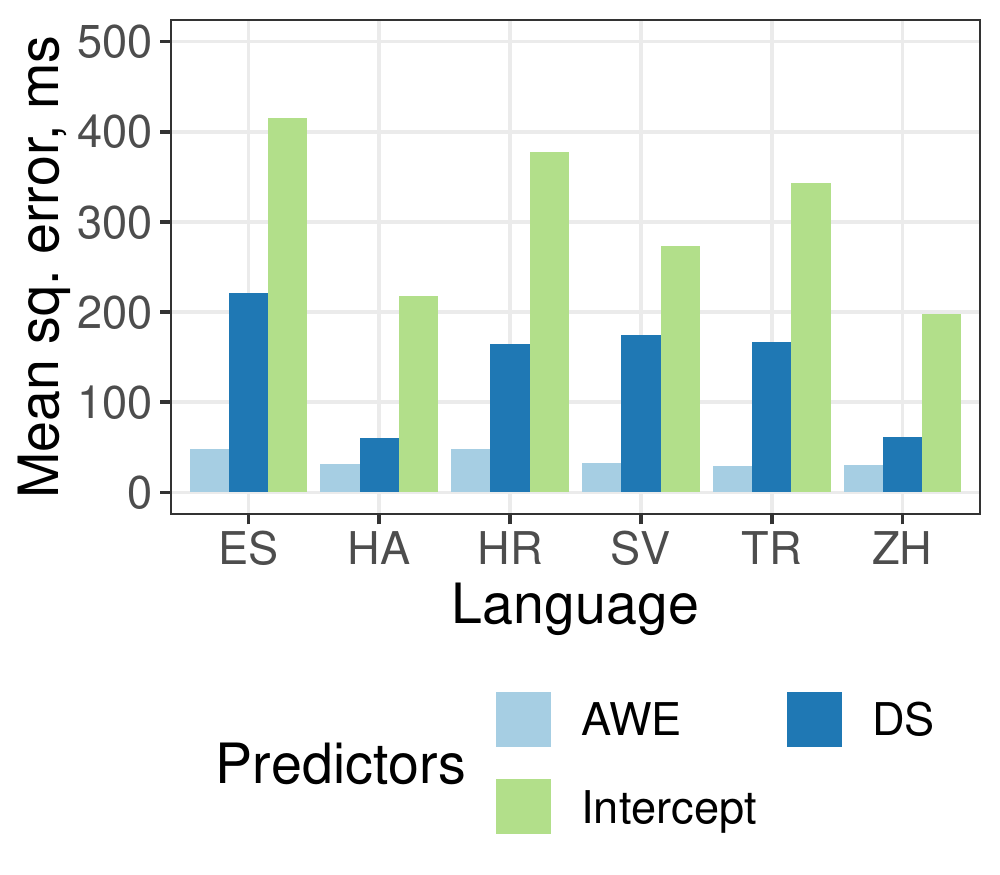}
        \captionof{figure}{Mean squared error of linear models predicting absolute word duration.}
        \label{fig:durreg}
    \end{minipage}\hfill
    \begin{minipage}{0.32\textwidth}
        \centering
        \includegraphics[width=\linewidth]{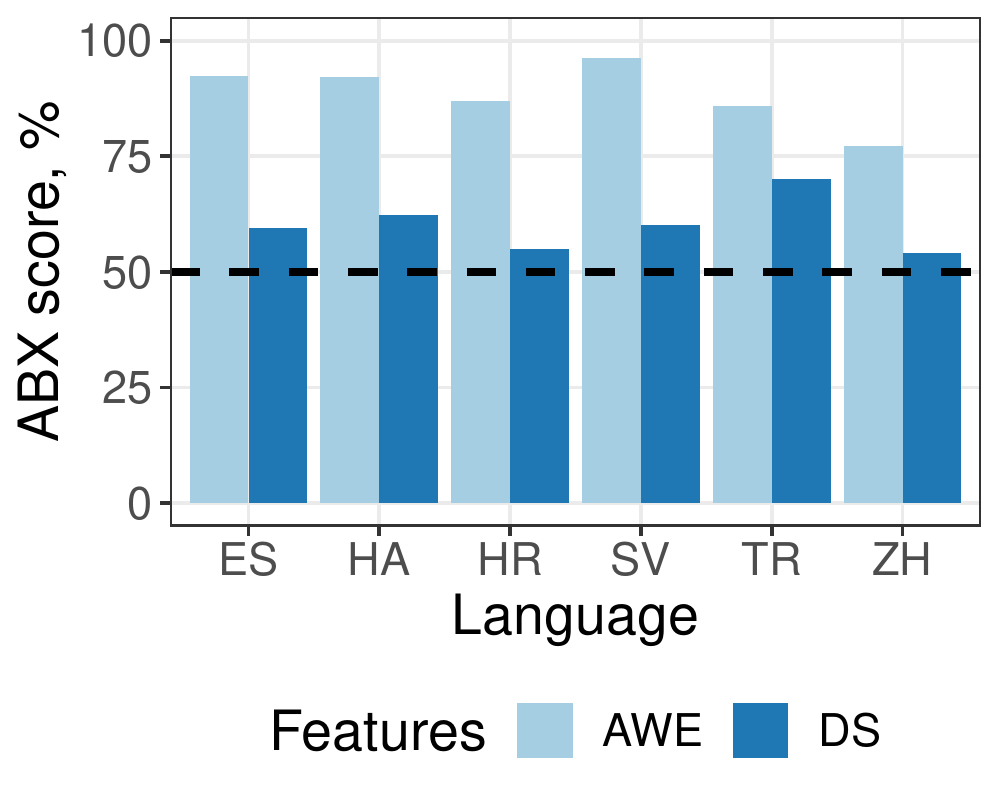}
        \caption{ABX scores in the task with words A and X matched on word duration, and B and X on speaker identity.}
        \label{fig:spkdurabx}
    \end{minipage}\hfill
    % \begin{minipage}{0.32\textwidth}
    %     \centering
    %     \includegraphics[width=\linewidth]{figures_old/ha_duration_vs_distance.pdf}
    %     \caption{Absolute difference in duration against distance in AWEs for Hausa, with linear regression models fitted to the data. \ym{Update plot.}}
    %     \label{fig:duration_vs_distance}
    % \end{minipage}
 \vspace{-10pt}
\end{figure}

\textbf{Speaker identity.} First, we look at whether the AWEs preserve any information about speaker identity,
despite being trained to ignore the variation across speakers.
We train a multiclass logistic regression classifier on $80\%$ of the embedded words to predict the speaker identity, and then test it on the held-out $20\%$ of words. Figure~\ref{fig:spkid} shows that the learned AWEs predict speaker identity worse than the DS baseline, but better than the majority class baseline:
that is, they abstract away from speaker characteristics to some degree, but not completely.
%In other words, the AWEs encode some information about the speakers' characteristics, although less than the baseline obtained by downsampling the data directly, once again confirming that the CAE-RNN successfully learns to abstracts away from such characteristics.

\textbf{Word duration.} Next, we test whether the
fixed-dimensional
AWEs preserve information about a basic acoustic property of a word---its absolute duration in milliseconds---without such information being explicitly provided. 
%despite their fixed dimensionality.
% Note that while the duration of word depends on its type (i.e., how many phones there are in the word), the same word can be pronounced fast or slowly, so the word type is not predictive of its absolute duration.
For each language, we train a linear regression model on $80\%$ of the embedded words to predict the word's absolute duration, and then test the model on the held-out $20\%$ of the words. Figure~\ref{fig:durreg} shows that the learned AWEs predict the word duration better than the DS baseline and the intercept baseline (i.e., a linear regression that only fits an intercept, thus always predicting the mean duration),
with $R^2$ in the range $0.85$--$0.91$ (not shown in Figure~\ref{fig:durreg}).
This suggests that the AWEs successfully encode temporal sequences into fixed-dimensional vectors while preserving information about their length.
%However, a word's absolute duration confounds acoustic properties of that word as a speech segment (random variation in the speech rate) and its phonetic properties (number of phones).
However, a word's absolute duration reflects not only random variation in the speech rate (i.e., duration as an acoustic property of the word as a speech segment), but also the number of phones in the word (i.e., the word's phonetic properties): \textit{category} on average takes longer to say than \textit{cat}.
To consider the duration as a purely acoustic property, we next look at various instances of the same word.

\textbf{Segment duration vs.\ speaker identity.}
We know that our AWEs encode some information about both segment duration and speaker identity,
% In principle, the AWEs should abstract away from both random acoustic variation (e.g. of signals (and focus on encoding phonetic information),
but do they encode both kinds of signal equally well? To test this, we design an ABX task with three instances of the same word, where A and X are generated by different speakers (but have similar duration, within a factor of $1.1$), while B and X are generated by the same speaker (but are different in duration by a factor of at least $1.5$). A score higher than $50\%$ indicates that word duration is encoded to a higher degree in the embedding space, while a score lower than $50\%$ shows that speaker identity is encoded better. Figure~\ref{fig:spkdurabx} shows that, while the DS baseline performs nearly at chance for $4$ out of $6$ languages, in our AWEs the absolute duration is a more distinctive feature than the speaker identity. Note that in this case there are no \textit{phonetic} differences between the acoustic words, which suggests the segment's duration is encoded in the AWEs as an \textit{acoustic} feature. Next, we test whether the AWEs also encode phonetic information.

% \subsection{Phonetic similarity and speaker characteristics}

% \subsection{Phonetic characteristics}

\begin{figure}
\centering
    \begin{minipage}{0.49\textwidth}
        \centering
        % \includegraphics[width=\linewidth]{figures/len_classif_2col.pdf}
        % \caption{Classification accuracy of models predicting number of phones in a word.}
        \includegraphics[width=\linewidth]{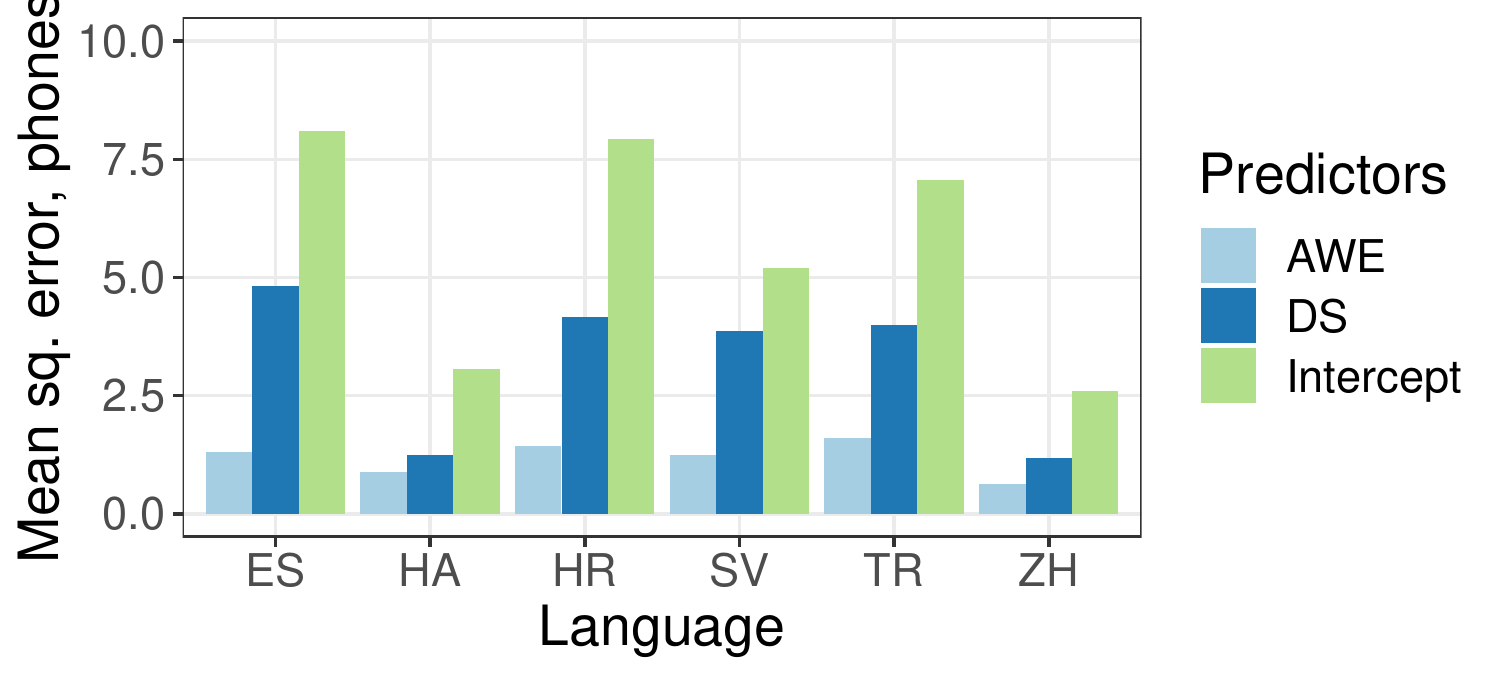}
        \caption{Mean squared error of linear models predicting number of phones in a word.}        
        % \\\hspace{\textwidth}
        \label{fig:lenclass}
    \end{minipage}\hfill
    \begin{minipage}{0.49\textwidth}
    \includegraphics[width=\textwidth]{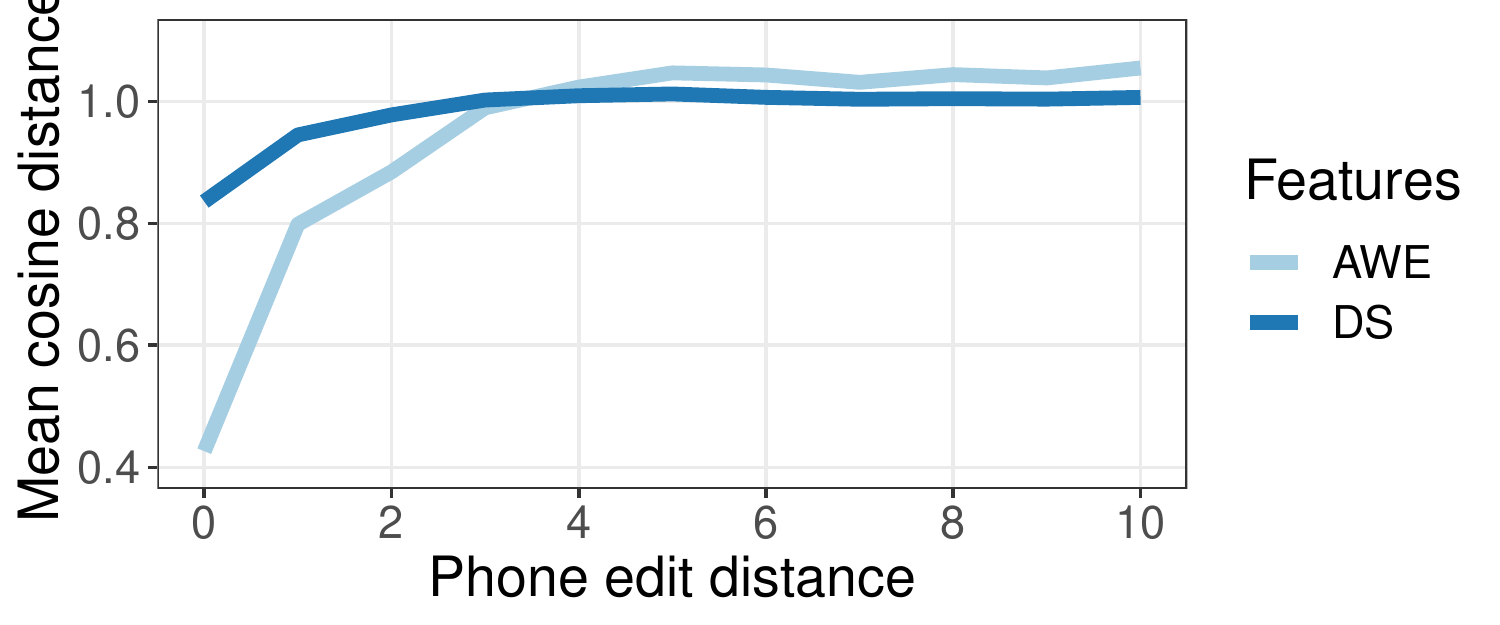}
    \caption{Phone edit distance between pairs of Hausa acoustic words against average cosine distance between their representations.}
    \label{fig:edit_distances}
    \end{minipage}\hfill
    \centering
    % \begin{minipage}{0.32\textwidth}
    %      \centering
    %      \includegraphics[width=\linewidth]{figures/spk_abx.pdf}
    %     \caption{Discrimination scores in a speaker-identity ABX task.}
    %     \label{fig:spkabx}
    % \end{minipage}
    % \begin{minipage}{0.32\textwidth}
    %      \centering
    %  \includegraphics[width=\linewidth]{figures_old/ha_speaker_boxplot.pdf}
    %  \caption{Distances between Hausa word embeddings spoken by the same and different speakers for different monolingual models. Only word pairs of the same type were considered.}
    %  \label{fig:speaker_boxplot}
    %  \end{minipage}
 \vspace{-10pt}
\end{figure}

\textbf{Number of phones.} To see how well our AWEs encode linguistically meaningful information, we look at the properties related to the words' phonetic content. First, we test whether the AWEs encode the information about the number of phones in a word.
%We train/test a multiclass logistic regression classifier to predict the number of phones in the words, using a $80\%$--$20\%$ split, as before. Figure~\ref{fig:lenclass} shows that the AWEs predict the number of phones better than both a logistic classifier trained on the DS data and a majority class classifier, suggesting that the AWE encode some information about the number of phones.
We train/test a linear regression model to predict the number of phones in the words, using an $80$/$20\%$ split, as before. Figure~\ref{fig:lenclass} shows that the AWEs predict the number of phones better than both the DS data and the intercept baseline (i.e., a linear regression always predicting the mean value), with $R^2$ in the range $0.71$--$0.84$ (not shown in Figure~\ref{fig:lenclass}), suggesting that the AWE encode some information about the number of phones.
% However, the high classification accuracy may be a by-product of the effective encoding of acoustic duration in the AWEs, the confound we discussed earlier.

% This suggests our AWEs encode some information about the words' phonetic properties. 
% To illustrate this result, Figure~\ref{fig:duration_vs_distance} shows a scatter plot of the absolute duration difference between pairs of Hausa words against the cosine distance between their AWEs (similar trends are obtained for the other languages). While the duration difference does not give a clear indication of the distance between their DS AWEs, words of similar duration tend to be closer in the CAE-RNN AWEs. Together, these results show that CAE-RNN AWEs do encode information about basic sequential properties of acoustic words, such as their duration and the number of phones.

% Acoustic words implicitly encode some speaker characteristics. A model of AWEs, however, needs to abstract away from such characteristics and instead organize its embedding space to efficiently encode words' phonetic properties:
\textbf{Phonetic similarity.} If our AWEs also encode words' phonetic properties, we expect phonetically similar words to be closer in the embedding space than dissimilar words.
% , and word type to be a more important feature of the space than speaker identity.
To test this, we look at whether the cosine distance between pairs of AWEs increases with the phone edit distance between the words (i.e., phonetic dissimilarity). Figure~\ref{fig:edit_distances} shows the results for Hausa (the trends are similar in the other languages): we observe the expected trend both in the DS baseline and in the AWEs, but in the AWEs
words that are more phonetically similar have more similar representations compared to the DS
% the cosine distance increases faster with the edit distance between the words
(which is especially evident for the pairs with edit distance zero: instances of the same word and/or homophones). This confirms that our AWEs encode some of the words' phonetic properties. 
% This shows that phonetically similar words tend to cluster together in the AWE space.
%For all three models, the separation between words increase systematically as we consider pairs differing in 1, 2, etc.\ phones.

 \begin{figure}
    \centering
    \begin{minipage}{0.49\textwidth}
        \centering
        \includegraphics[width=\linewidth]{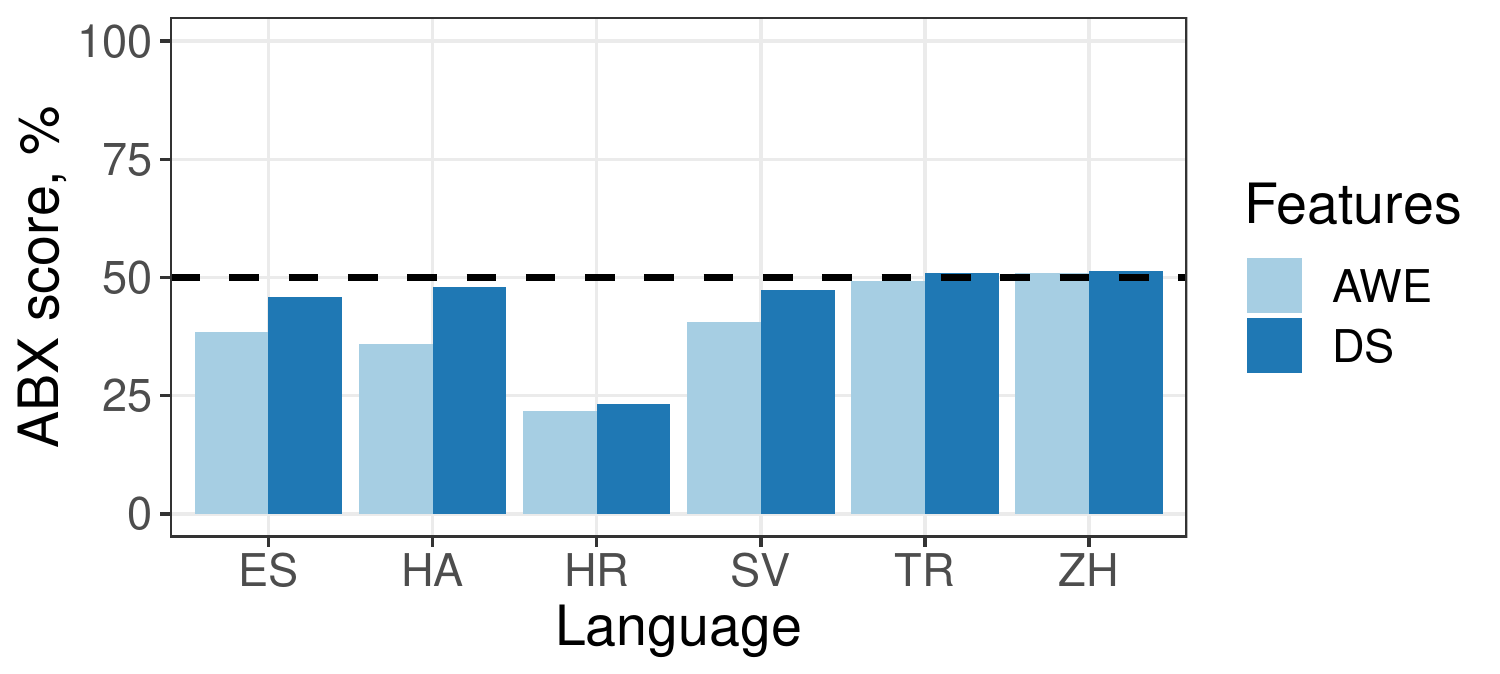}
        \caption{ABX scores in the task where words A and X differ in their initial phone, and B and X in another phone.}
        \label{fig:posabx}
    \end{minipage}\hfill
    \begin{minipage}{0.49\textwidth}
        \centering
        \includegraphics[width=\textwidth]{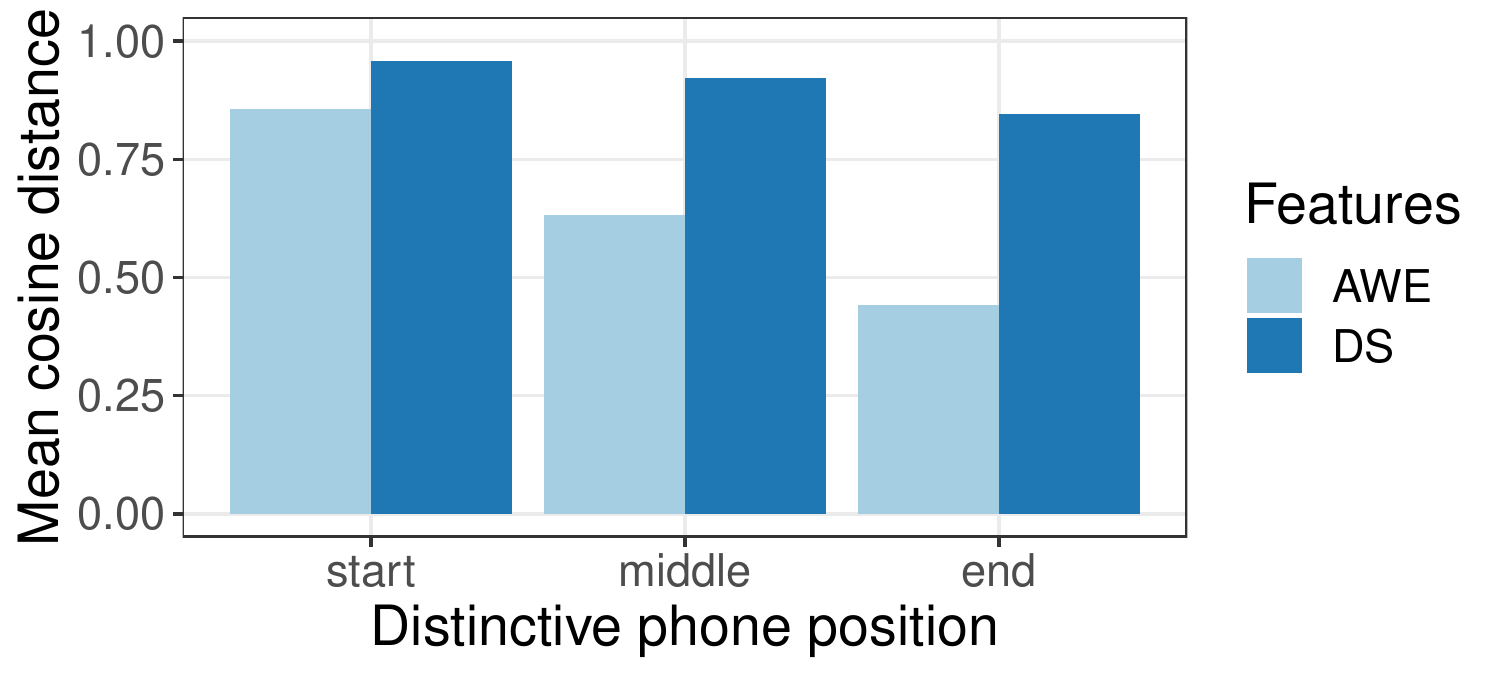}
        \caption{Average cosine distances between pairs of Hausa AWEs for words differing in one phone, depending on the position of that phone.}
        \label{fig:edit_positions_boxplot}
    \end{minipage}\hfill
% \vspace{-15pt}    
\end{figure}

% \subsection{Word onset bias}

\textbf{Word onset bias.} Finally, we ask whether the AWEs exhibit the human-like word onset bias: considering the first sound of the word more `prominent' than its other sounds.
% Under this bias we understand a broad range of effects showing that human speakers give more importance to the initial sound of the word (compared to its other sounds): speakers strengthen the initial sound in the articulation \citep{fougeron1997, keating1999}, listeners can capture the distinctions between word-initial and word-final sounds \citep{shatzman2006}, word-initial sounds have a special status in spoken word recognition \citep{marslen1989}, the first letter in a word is a more efficient cue for lexical retrieval than other letters \citep{brown1990}, etc.
We use an ABX task and a comparison of distances between words, in both methods focusing on pairs of words with phone edit distance of $1$. In the ABX task, words A and X differ in their first phone, while B and X differ in another phone (e.g., X: \textit{take}, A: \textit{\textbf{c}ake}, B: \textit{ta\textbf{p}e}). A score of $50\%$ indicates no difference depending on the distinctive phone position (i.e., X is equally close to A and B), a score below $50\%$ indicates the expected bias (i.e., X is closer to A than to B), and a score above $50\%$ indicates a bias in the opposite direction.
% A: \textit{lake}, B: \textit{cage}, X: \textit{cake}.
%\footnote{Ideally, the phone contrast should be the same in both pairs, A--X and B--X, as the [k--l] in A:~\textit{lake}, B:~\textit{kale}, X:~\textit{cake}. This means, however, that the X word should have two identical phones, with one of them occurring at the beginning of the word, and that the corresponding minimal pairs A and B exist. There are almost no triplets satisfying these requirements in the GlobalPhone dev data, so we consider triplets where the contrasting phones in A--X and B--X differ.}
Figure~\ref{fig:posabx} shows that the AWEs score below $50\%$ in most languages, indicating a larger distance between words that differ in their first phone compared to words that differ in another phone, which corresponds to the predicted bias. Importantly, the scores are lower in the AWEs than in the DS data, suggesting that this bias does not completely arise from the data alone,
but is learned by the model
(although the presence of the bias in the DS data suggests that the first phone may provide a stronger signal---e.g., in terms of duration---than other phones).
In addition, when we look at the distances between pairs of AWEs for words that differ in a single phone in Hausa (Figure~\ref{fig:edit_positions_boxplot}, with similar results in other languages), we observe larger distances when the distinctive phone is at the beginning of the word (rather than in the middle or at the end), and this tendency is stronger in the AWEs than in the DS data, in line with the results of our ABX task.

\section{Conclusion}

We presented an analysis of basic properties of a particular type of acoustic word embeddings, which are based on an encoder-decoder model. We showed that these embeddings can succeed in encoding some characteristics of the words' phonetic content, yet they also preserve information about an acoustic word's absolute duration and speaker identity.
%(1) encode information about the duration of a word and the number of phones in it, (2) abstract away from speaker characteristics and shape around word phonetic similarity and word identity, and (3) exhibit the human-like word onset bias.
We also show that AWEs can exhibit a bias towards treating the first sound in the word as a more important part of the signal, compared to the other sounds---a pattern mirroring the empirical data observed in human speakers. These results suggest that AWEs show some promise as a modeling tool in cognitive science, and encourage further research in this direction. AWEs can provide a straightforward connection between human speech processing and lexical storage and access, as acoustic words of any duration are situated within a feature space that is easy to probe with various tests such as the ones presented in this study. While AWEs are devoid of any semantics, they could be combined with speech-based \citep{chung2018a} or textual semantic word embeddings \citep[as in][]{chen2018}, potentially informing more accurate models of human lexical memory and access, which need to take into account word pronunciations or their acoustic properties \citep{aydelott2004, andruski1994}.

\section*{Acknowledgements}

This work is based on research supported in part by the National Research Foundation of South Africa (grant number: 120409), a James S. McDonnell Foundation Scholar Award (220020374), an ESRC-SBE award (ES/R006660/1), and a Google Faculty Award for HK. We thank Kate McCurdy, Adam Lopez, Ramon Sanabria and other members of the AGORA group at the Edinburgh School of Informatics for their valuable feedback.

\bibliography{references}
\bibliographystyle{iclr2020_conference}

\appendix
\section{Appendix}

\textbf{Model training.} We train six monolingual CAE-RNN models (one per language) on data extracted from GlobalPhone, a non-parallel corpus of read newspaper articles \citep{schultz2002}. Each language has $16$ hours of training and $2$ hours of test data on average, with test data sampled from held-out speakers. To prepare word pairs for training the model, we first create a list of all words in the training data (obtained through forced word alignments) with duration of at least $500$ ms and containing at least $5$ phones. We then randomly pair words of the same type to create $100,000$ pairs. Following \citet[][]{kamper2020}, we pre-train the model as an autoencoder for $15$ epochs to initialize its parameters, and then train it for $25$ epochs using the existing architecture: $3$ hidden layers ($400$ gated recurrent units each) in both the decoder and encoder, and an embedding dimensionality of $130$. For reference, on the `same-different' task, these models score $60$--$85\%$, depending on the language.

\begin{table}[H]
\centering
    % \begin{tabular}{p{0.4cm}p{1.1cm}p{3.8cm}p{1.3cm}p{1.3cm}}
    \begin{tabular}{lllrr}
    \hline
         Code & Language & Family (branch) & Test speakers & Phones per word: mean (SD) \\
    \hline
         ES & Spanish & Indo-European (Romance) & $10$ & $4.9$ ($3.1$)\\
         HA & Hausa & Afroasiatic (Chadic) & $10$ & $4.2$ ($1.8$) \\
         HR & Croatian & Indo-European (Slavic) & $10$ & $5.4$ ($2.8$)\\
         SV & Swedish & Indo-European (Germanic) & $9$ & $4.1$ ($2.3$) \\
         TR & Turkish & Turkic (Oghuz) & $11$ & $6.0$ ($2.7$) \\
         ZH & Mandarin & Sino-Tibetan (Mandarin) & $11$ & $3.6$ ($1.6$) \\
    \hline
        \end{tabular}
    \captionof{table}{Characteristics of the test data.}
    \label{tab:data}
%     \vspace{-6mm}
\end{table}

\end{document}

%% file: math_commands.tex
\usepackage{amsmath,amsfonts,bm}

\def\eqref#1{equation~\ref{#1}}
\def\1{\bm{1}}

\DeclareMathAlphabet{\mathsfit}{\encodingdefault}{\sfdefault}{m}{sl}
\SetMathAlphabet{\mathsfit}{bold}{\encodingdefault}{\sfdefault}{bx}{n}

%% file: main.bbl
\begin{thebibliography}{25}
\providecommand{\natexlab}[1]{#1}
\providecommand{\url}[1]{\texttt{#1}}
\expandafter\ifx\csname urlstyle\endcsname\relax
  \providecommand{\doi}[1]{doi: #1}\else
  \providecommand{\doi}{doi: \begingroup \urlstyle{rm}\Url}\fi

\bibitem[Andruski et~al.(1994)Andruski, Blumstein, and Burton]{andruski1994}
Jean~E. Andruski, Sheila~E. Blumstein, and Martha Burton.
\newblock The effect of subphonetic differences on lexical access.
\newblock \emph{Cognition}, 52:\penalty0 163--187, 1994.

\bibitem[Aydelott \& Bates(2004)Aydelott and Bates]{aydelott2004}
Jennifer Aydelott and Elizabeth Bates.
\newblock Effects of acoustic distortion and semantic context on lexical
  access.
\newblock \emph{Language and Cognitive Processes}, 19:\penalty0 29--56, 2004.

\bibitem[Brown \& Knight(1990)Brown and Knight]{brown1990}
Alan~S. Brown and Kevin~K. Knight.
\newblock Letter cues as retrieval aids in semantic memory.
\newblock \emph{The American Journal of Psychology}, pp.\  101--113, 1990.

\bibitem[Chen et~al.(2018)Chen, Huang, Shen, Lee, and Lee]{chen2018}
Yi-Chen Chen, Sung-Feng Huang, Chia-Hao Shen, Hung-Yi Lee, and Lin-Shan Lee.
\newblock Phonetic-and-semantic embedding of spoken words with applications in
  spoken content retrieval.
\newblock In \emph{2018 IEEE Spoken Language Technology Workshop (SLT)}, pp.\
  941--948, 2018.

\bibitem[Chung \& Glass(2018)Chung and Glass]{chung2018a}
Yu-An Chung and James Glass.
\newblock Speech2vec: A sequence-to-sequence framework for learning word
  embeddings from speech.
\newblock In \emph{Proceedings of Interspeech}, pp.\  811--815, 2018.

\bibitem[Chung et~al.(2016)Chung, Wu, Shen, Lee, and Lee]{chung2016}
Yu-An Chung, Chao-Chung Wu, Chia-Hao Shen, Hung-Yi Lee, and Lin-Shan Lee.
\newblock Audio word2vec: Unsupervised learning of audio segment
  representations using sequence-to-sequence autoencoder.
\newblock In \emph{Proceedings of Interspeech}, pp.\  765--769, 2016.

\bibitem[Chung et~al.(2018)Chung, Weng, Tong, and Glass]{chung2018b}
Yu-An Chung, Wei-Hung Weng, Schrasing Tong, and James Glass.
\newblock Unsupervised cross-modal alignment of speech and text embedding
  spaces.
\newblock In \emph{Advances in Neural Information Processing Systems}, pp.\
  7354--7364, 2018.

\bibitem[Feldman et~al.(2013)Feldman, Griffiths, Goldwater, and
  Morgan]{feldman2013}
Naomi~H. Feldman, Thomas~L. Griffiths, Sharon Goldwater, and James~L. Morgan.
\newblock A role for the developing lexicon in phonetic category acquisition.
\newblock \emph{Psychological Review}, 120:\penalty0 751--778, 2013.

\bibitem[Fougeron \& Keating(1997)Fougeron and Keating]{fougeron1997}
C{\'e}cile Fougeron and Patricia~A. Keating.
\newblock Articulatory strengthening at edges of prosodic domains.
\newblock \emph{The Journal of the Acoustical Society of America},
  101:\penalty0 3728--3740, 1997.

\bibitem[Ghannay et~al.(2016)Ghannay, Est{\`e}ve, Camelin, and
  Del{\'e}glise]{ghannay2016}
Sahar Ghannay, Yannick Est{\`e}ve, Nathalie Camelin, and Paul Del{\'e}glise.
\newblock Evaluation of acoustic word embeddings.
\newblock In \emph{Proceedings of the 1st Workshop on Evaluating Vector-Space
  Representations for NLP}, pp.\  62--66, 2016.

\bibitem[Grand et~al.(2018)Grand, Blank, Pereira, and Fedorenko]{grand2018}
Gabriel Grand, Idan~Asher Blank, Francisco Pereira, and Evelina Fedorenko.
\newblock Semantic projection: Recovering human knowledge of multiple, distinct
  object features from word embeddings.
\newblock \emph{arXiv:1802.01241}, 2018.

\bibitem[Holzenberger et~al.(2018)Holzenberger, Du, Karadayi, Riad, and
  Dupoux]{holzenberger2018}
Nils Holzenberger, Mingxing Du, Julien Karadayi, Rachid Riad, and Emmanuel
  Dupoux.
\newblock Learning word embeddings: Unsupervised methods for fixed-size
  representations of variable-length speech segments.
\newblock In \emph{Proceedings of Interspeech}, pp.\  2683--2687, 2018.

\bibitem[Jusczyk \& Aslin(1995)Jusczyk and Aslin]{jus95a}
Peter~W. Jusczyk and Richard~N. Aslin.
\newblock Infants' detection of the sound patterns of words in fluent speech.
\newblock \emph{Cognitive Psychology}, 29:\penalty0 1--23, 1995.

\bibitem[Jusczyk et~al.(1999)Jusczyk, Houston, and Newsome]{jus99b}
Peter~W. Jusczyk, Derek~M. Houston, and Mary Newsome.
\newblock The beginnings of word segmentation in {E}nglish-learning infants.
\newblock \emph{Cognitive Psychology}, 39:\penalty0 159--207, 1999.

\bibitem[Kamper(2019)]{kamper2019}
Herman Kamper.
\newblock Truly unsupervised acoustic word embeddings using weak top-down
  constraints in encoder-decoder models.
\newblock In \emph{Proceedings of ICASSP}, pp.\  6535--3539, 2019.

\bibitem[Kamper et~al.(2020)Kamper, Matusevych, and Goldwater]{kamper2020}
Herman Kamper, Yevgen Matusevych, and Sharon Goldwater.
\newblock Multilingual acoustic word embedding models for processing
  zero-resource languages.
\newblock \emph{arXiv:2002.02109}, 2020.

\bibitem[Keating et~al.(1999)Keating, Wright, and Zhang]{keating1999}
Patricia Keating, Richard Wright, and Jie Zhang.
\newblock Word-level asymmetries in consonant articulation.
\newblock \emph{UCLA Working Papers in Phonetics}, pp.\  157--173, 1999.

\bibitem[Levin et~al.(2013)Levin, Henry, Jansen, and Livescu]{levin2013}
Keith Levin, Katharine Henry, Aren Jansen, and Karen Livescu.
\newblock Fixed-dimensional acoustic embeddings of variable-length segments in
  low-resource settings.
\newblock In \emph{IEEE Workshop on Automatic Speech Recognition and
  Understanding}, pp.\  410--415, 2013.

\bibitem[Marslen-Wilson \& Zwitserlood(1989)Marslen-Wilson and
  Zwitserlood]{marslen1989}
William Marslen-Wilson and Pienie Zwitserlood.
\newblock Accessing spoken words: The importance of word onsets.
\newblock \emph{Journal of Experimental Psychology: Human Perception and
  Performance}, 15:\penalty0 576--585, 1989.

\bibitem[Matusevych et~al.(2020)Matusevych, Schatz, Kamper, Feldman, and
  Goldwater]{matusevych2020}
Yevgen Matusevych, Thomas Schatz, Herman Kamper, Naomi Feldman, and Sharon
  Goldwater.
\newblock Evaluating computational models of infant phonetic learning across
  languages.
\newblock \emph{Under review}, 2020.

\bibitem[Nematzadeh et~al.(2017)Nematzadeh, Meylan, and
  Griffiths]{nematzadeh2017}
Aida Nematzadeh, Stephan~C. Meylan, and Thomas~L. Griffiths.
\newblock Evaluating vector-space models of word representation, or, the
  unreasonable effectiveness of counting words near other words.
\newblock In \emph{Proceedings of CogSci}, pp.\  859--864, 2017.

\bibitem[Pereira et~al.(2016)Pereira, Gershman, Ritter, and
  Botvinick]{pereira2016}
Francisco Pereira, Samuel Gershman, Samuel Ritter, and Matthew Botvinick.
\newblock A comparative evaluation of off-the-shelf distributed semantic
  representations for modelling behavioural data.
\newblock \emph{Cognitive Neuropsychology}, 33:\penalty0 175--190, 2016.

\bibitem[Schatz et~al.(2013)Schatz, Peddinti, Bach, Jansen, Hermansky, and
  Dupoux]{schatz2013}
Thomas Schatz, Vijayaditya Peddinti, Francis Bach, Aren Jansen, Hynek
  Hermansky, and Emmanuel Dupoux.
\newblock {Evaluating speech features with the minimal-pair ABX task: Analysis
  of the classical MFC/PLP pipeline}.
\newblock In \emph{Proceedings of Interspeech}, pp.\  1781--1785, 2013.

\bibitem[Schultz(2002)]{schultz2002}
Tanja Schultz.
\newblock {GlobalPhone: A multilingual speech and text database developed at
  Karlsruhe University}.
\newblock In \emph{{Proceedings of ICSLP}}, pp.\  345--348, 2002.

\bibitem[Shatzman \& McQueen(2006)Shatzman and McQueen]{shatzman2006}
Keren~B. Shatzman and James~M. McQueen.
\newblock Segment duration as a cue to word boundaries in spoken-word
  recognition.
\newblock \emph{Perception \& Psychophysics}, 68:\penalty0 1--16, 2006.

\end{thebibliography}
